\pdfoutput=1
\documentclass[11pt]{article}

\usepackage[final]{acl}

\usepackage{times}
\usepackage{latexsym}
\usepackage[T1]{fontenc}
\usepackage[utf8]{inputenc}

\usepackage{microtype}

\usepackage{inconsolata}

\usepackage{graphicx}
\usepackage{hyperref}
\usepackage{url}
\usepackage{subcaption}
\usepackage{booktabs}
\usepackage{makecell}
\usepackage{amsmath}
\usepackage{amssymb}
\usepackage{enumitem}
\usepackage{graphicx}
\usepackage{lineno}
\usepackage{multirow}
\usepackage{multicol}
\usepackage{CJKutf8}
\usepackage{tcolorbox}
\usepackage{cuted}
\PassOptionsToPackage{table}{xcolor} % Ensure the [table] option is passed
\usepackage{xcolor} % Load the xcolor package

\title{Translationese-index: Using Likelihood Ratios \\ for Graded and Generalizable Measurement of Translationese}

\author{
 \textbf{Yikang Liu\textsuperscript{1}\thanks{Work done when Yikang Liu was an intern at Tongyi Lab. $^{\#}$Rui Wang and Hai Hu are corresponding authors.}},
 \textbf{Wanyang Zhang\textsuperscript{2}},
 \textbf{Yiming Wang\textsuperscript{1}},
 \textbf{Jialong Tang\textsuperscript{3}},
\\
 \textbf{Pei Zhang\textsuperscript{3}},
 \textbf{Baosong Yang\textsuperscript{3}},
 \textbf{Fei Huang\textsuperscript{3}},
 \textbf{Rui Wang\textsuperscript{1\#}},
 \textbf{Hai Hu\textsuperscript{4\#}}
\\
 \textsuperscript{1}Shanghai Jiao Tong University~~
 \textsuperscript{2}Peking University 
\\
 \textsuperscript{3}Tongyi Lab~~
 \textsuperscript{4}City University of Hong Kong
\\
 \small{
   \textbf{Correspondence:} \href{mailto:email@domain}{yikangliu@sjtu.edu.cn};
   \href{mailto:email@domain}{wangrui12@sjtu.edu.cn};
   \href{mailto:email@domain}{hu.hai@cityu.edu.hk}
 }
}

\begin{document}
\maketitle
\begin{abstract}
Translationese refers to linguistic properties that usually occur in translated texts. Previous works study translationese by framing it as a binary classification between original texts and translated texts. 
In this paper, we argue that translationese should be graded instead of binary and propose the first measure for translationese---the translationese-index (T-index), computed from the likelihood ratios of two contrastively fine-tuned language models (LMs). 
We use synthesized translations and translations in the wild to evaluate T-index's generalizability in cross-domain settings and its validity against human judgments.
Our results show that T-index can generalize to unseen genres, authors, and language pairs. Moreover, T-index computed using two 0.5B LMs fine-tuned on only 1-5k pairs of synthetic data can effectively capture translationese, as demonstrated by alignment with human pointwise ratings and pairwise judgments.
% The relative differences in T-index and the absolute values well predict human pairwise preferences and pointwise ratings, respectively. 
% We find that the relative differences in T-indices between translations can predict pairwise judgments obtained from human annotators; and the absolute values of T-indices correlate well with human ratings of degrees of translationese (Pearson's $r = 0.568$).
Additionally, the correlation between T-index and existing machine translation (MT) quality estimation (QE) metrics such as BLEU and COMET is low, suggesting that T-index is not covered by these metrics and
can serve as a complementary metric in MT QE.

\includegraphics[width=.8em,height=.8em]{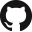}\hspace{.3em}\parbox{\dimexpr\linewidth-2\fboxsep-2\fboxrule}{\footnotesize \url{https://github.com/yikang0131/TranslationeseIndex}}
\end{abstract}

\section{Introduction}
\label{sec:intro}
\begin{figure*}[ht]
    \centering
    \includegraphics[width=\textwidth]{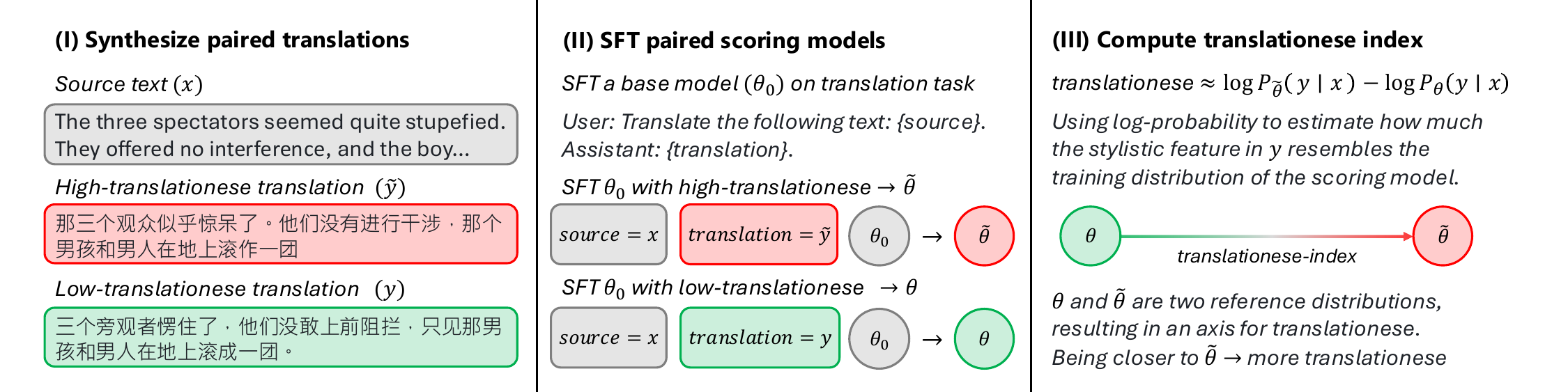}
    \caption{Illustration of the pipeline for translationese measuring using likelihood ratios of a pair of LMs.}
    \label{fig:intro:demo}
\end{figure*}

Translationese refers to linguistic properties that are often introduced in the translation process that are different from those of texts originally written in that language~\citep{gellerstam-1986-translationese}.
While such properties are not inherently undesirable, they often lead to unnatural and non-native-like language that differs from idiomatic and authentic texts. 

Translationese in translations, particularly machine translations (MTs), presents significant challenges in the era of large language models (LLMs). Many multilingual resources synthesized through MT have been reported as low quality~\citep{kreutzer-etal-2022-quality}. Models trained on these ``noisy'' MT datasets often struggle to generalize effectively in real-world tasks that do not involve translation~\citep{church-et-al-2025-trends}. This issue extends even to high-resource languages: for Chinese, the second most resource-rich language, LLMs frequently produce unnatural texts resembling translationese in monolingual natural language generation tasks~\citep{guo-etal-2024-llm-english}.

We believe this problem can be alleviated through a quantitative measure of translationese that enables selection of the most authentic and natural translation from multiple MT outputs. However, previous attempts to detect or measure translationese have failed to meet this goal. Existing approaches either develop binary classifiers to distinguish between translated and original texts~\citep{baroni-and-bernardini-2005-translationese,koppel-2011-translationese,volansky-et-al-2015-features,hu-and-kübler-2021-translated,pylypenko-etal-2021-comparing}, or rely on distributional statistics computed across batches of texts rather than individual samples~\citep{freitag-etal-2022-natural,guo-etal-2024-llm-english,li-etal-2025-literalism,flamich-etal-2025-cannot}, in which original texts are regarded as no-translationese distribution. The former approach lacks continuous measurement capabilities, while the latter cannot score individual translations.

In this paper, we want to find a graded and generalizable measurement of translationese. To this end, we first reframe how translationese should be discovered. We argue that the degree of translationese should be directly compared among translations. This new framework offers two key benefits: first, graded comparisons can be more easily observed, and second, it isolates confounding factors that arise in classification between original and translated texts, thereby enabling better generalization.

% With the new problem formulation, we propose \textit{translationese index} (T-index), the first method for graded and generalizable measurement of translationese, based on likelihood ratios of two contrastively fine-tuned LLMs, a low-translationese model and a high-translationese model. 
% The method contrasts the distance between the samples and the training distribution of each model, estimating the level of translationese.

Under the new problem formulation, we propose \textit{translationse index} (T-index) and compare it with several supervised and unsupervised baselines on synthetic and human-annotated data.
We have several major findings: 
(1) The proposed method is generalizable to multiple genres, authors, and language pairs, even when the backbone LMs are fine-tuned with only 1k synthesized samples. 
(1) T-index is correlated with both pointwise (Pearson's $r=0.418$) and pairwise evaluations of translationese by expert human annotators. 
(3) T-index has very low correlation with existing MT quality estimation (QE) metrics, suggesting that it might be a novel aspect not yet covered by existing metrics.

% TO conclusion, \blue{which can be especially useful in the era of Large Language Model as existing MT QE measures have been found to be increasingly difficult in distinguishing MTs of different qualities CITE}.

The paper is organized as follows. We first introduce our new problem formulation and T-index in \S\ref{sec:t-index}. Next, we describe the multi-genre synthetic benchmark used for translationese measurement in \S\ref{sec:benchmark}.
In \S\ref{sec:generalizable}, we evaluate the cross-domain generalization of T-index and several baselines on the synthetic benchmark.
Finally, we evaluate T-index against human annotations for MTs in the wild in \S\ref{sec:graded}.

\section{T-index: using likelihood ratios to measure translationese}
\label{sec:t-index}

\paragraph{Problem formulation.} 
Unlike previous research, which was framed as a binary text-classification task between translated and non-translated text, 
we compare different translations of the same source text, attempting to provide a continuous measure of translationese. 
The new framing is motivated by the following rationales: 
\begin{enumerate}[leftmargin=*]
    \setlength{\itemsep}{0pt}
    \item Whether the texts are translated or not does not determine the degree of translationese, i.e., translated texts can also be authentic, and non-translated ones might be unnatural.
    \item The binary-classification between translated and non-translated texts is sometimes confounded by features unrelated to translationese, such as the topic of the texts~\citep{amponsah-kaakyire-etal-2022-explaining,borah-etal-2023-measuring}, hindering the discovery of translationese-specific features. 
    % Translated and non-translated texts essentially come from different sources, which might hinder the discovery of genuine linguistic features related to translationese.
\end{enumerate}

In the new formulation, we do not consider non-translated texts as having ``no translationese'', but instead aim to measure the degree of translationese of each individual translation.

Specifically, for a source text $x$ and its translation $y$, we want to estimate the degree of translationese in $y$. 
The goal is to find a proper scoring function $f(x,y,\theta)$, parameterized by $\theta$. For each sample $(x,y)$, $f$ will yield a score that directly predicts the degree of translationese. 

% As reported in the classification between original and translated texts, the supervised classifiers might learn confounding factors that are not necessarily related to translationese, such as the topic of the texts~\citep{amponsah-kaakyire-etal-2022-explaining}. Such features might hinder the cross-domain generalization. 

% We assume that the scoring function can perform cross-domain generalization well if it indeed captures abstract translationese-related features. 
% Therefore, we set two confounding variables: (1) \textit{\textbf{genre}} and (2) \textit{\textbf{author}}, which means that the test samples might be sourced from different genres or translated by different authors. The automated translationese measurement is expected to tease out the translationese features from the stylistic variance brought by the genre and author.

In order to make the scoring function genuinely generalizable, we consider two possible confounding variables: (1) \textit{\textbf{genre}} and (2) \textit{\textbf{author}}. 
Ideally, our scoring function should only capture abstract translationese-specific features, rather than genre- or author-related textual features of the translations. 
Thus, to test whether the scoring function is generalizable and robust, we include test samples from different genres or translated by different authors.

Suppose that we have a set of authors $\mathcal{A} = \{a_1, a_2, \ldots, a_n\}$, a set of genres $\mathcal{G} = \{g_1, g_2, \ldots, g_m\}$, we can denote a dataset as $\mathcal{D}_{g_i, a_i}$, where the source text is sampled from the genre $g_i$, and the translation is produced by the author $a_i$. $\mathcal{D}$ is a paired dataset, each sample containing a translation with a higher degree of translationese ($\Tilde{y}$) and a translation with a lower degree of translationese ($y$). The goal can be formalized as follows:

\begin{equation*}
    \max_{f} \sum_{(x, y, \Tilde{y}) \in \mathcal{D}}\mathbb{I}\left[f(x, \Tilde{y}, \theta) > f(x, y, \theta)\right].
\end{equation*}

\paragraph{Likelihood ratios as translationese index.} 
Inspired by the success of Likelihood Ratios (LLR) in OOD detection~\citep{ren-etal-2019-likelihood}, we propose to use LLR to measure translationese. Assume that the translation $y$ can be decomposed into three independent parts $\{\mathbf{y}_g, \mathbf{y}_a, \mathbf{y}_t\}$ as genre component, author component, and translationese component, resulting $\log \mathrm{P}_\theta(y \mid x) = \log\left[P_\theta(\mathbf{y}_g) P_\theta(\mathbf{y}_a) P_\theta(\mathbf{y}_t)\right]$. Given a paired dataset $\mathcal{D}$, we can contrastively fine-tune two scoring models $\theta$ (on low-translationese samples) and $\Tilde{\theta}$ (on high-translationese samples) (see Figure~\ref{fig:intro:demo}). T-index can be formalized as: 

\begin{align}
    \mathrm{T\mbox{-}index}(y \mid x) &= \log\frac{\mathrm{P}_{\Tilde{\theta}}(\mathbf{y}_g) \mathrm{P}_{\Tilde{\theta}}(\mathbf{y}_a) \mathrm{P}_{\Tilde{\theta}}(\mathbf{y}_t)}{\mathrm{P}_\theta(\mathbf{y}_g) \mathrm{P}_\theta(\mathbf{y}_a) \mathrm{P}_\theta(\mathbf{y}_t)} \notag \\
    &\approx \log\mathrm{P}_{\Tilde{\theta}}(\mathbf{y}_t) - \log\mathrm{P}_{\theta}(\mathbf{y}_t). \notag
\end{align}

Intuitively, if a translation $y$ is more likely to be high-translationese, the likelihood of $y$ given by the low-translationese model $\theta$ should be lower than that given by the high-translationese model $\Tilde{\theta}$. 
Since the other two components, genre and author, are shared between the two scoring models, we expect them to be canceled out. 
We also expect LLR to be robust in cross-domain generalization, because the shift in genre and author in testing samples should be captured by both models. 
We provide empirical confirmation of these assumptions in Appendix~\ref{sec:empirical-confirm}.

\paragraph{Roadmap for the validation of T-index.} 

We show the validity of T-index in two steps. 
In the first step, 
% we test the extreme case where a classifier needs to distinguish texts with very low translationese and texts with very high translationese. 
we formulate the problem as classification, 
% we follow previous work to frame the problem as classification, 
but instead of having translated and non-translated texts as the classes, we generate two classes of texts with extreme degrees of translationese, using carefully controlled prompts: one with very low translationese, the other with very high translationese, and verify that human annotators agree with the low- vs.~high-translationese distinction (\S\ref{sec:benchmark}). 
We then use T-index to perform text classification to examine the discriminative power and generalizability of T-index for texts on the two ends of translationese
(\S\ref{sec:generalizable}). 
% The idea is to examine if T-index can reliably distinguish low and high translationese. 
The second step uses real-world translations and abandons the text-classification paradigm (\S\ref{sec:graded}). We first ask human annotators to rate the degree of translationese in the sampled translations. We then show that T-index is in line with human ratings using various methods. 

\section{Constructing a synthetic benchmark of translationese}
\label{sec:benchmark}
% \section{Constructing a toy benchmark and collecting human judgments}

% Different from the classification between the original and translated texts, the graded measurement of translationese lacks inherent ground-truth labels, relying solely on human reading feedback. However, it is costly to collect human judgments concerning translationese at scale, especially when it is not verified whether humans can measure translationese reliably, as previous works show that it is a difficult task for humans (even professional translators) to distinguish translated and non-translated texts~\citep{baroni-and-bernardini-2005-translationese,wein-2023-human}.

% We choose to synthesize a benchmark contrasting low-translationese and high-translationese translations of the same set of multi-genre texts (\S\ref{sec:toy:benchmark}). 
% The synthetic data might lack diversity as a trade-off, but it is well-controlled with obvious divergence in the two types of translations, where significant differences in corpus-level linguistic features are observed (\S\ref{sec:toy:dataset-stat}). 
% This dataset serves as the starting point to test both manual (\S\ref{sec:toy:human:judgments}) and automatic evaluation in an easier experimental setting.

In this section, we describe how we construct a synthetic dataset containing low-translationese and high-translationese of the same set of source texts (\S\ref{sec:benchmark:data-gen}). 
We then present corpus-level statistics of the dataset, demonstrating the difference in linguistic features of the low- and high-translationese texts (\S\ref{sec:benchmark:data-gen}). 
Finally, we conduct human annotation to demonstrate that humans can indeed capture these differences in degrees of translationese (\S\ref{sec:benchmark:annotation}).

\subsection{Data generation}
\label{sec:benchmark:data-gen}

First, we sample English texts from 7 varied sources (\textbf{\textit{genre}}), including three 19th-century novels written by Charles Dickens and samples of four genres in \textit{The Corpus of Contemporary American English} (COCA; ~\citealp{davies-2008-coca}). The three novels include \textit{Oliver Twist} (1838), \textit{Great Expectations} (1861), and \textit{A Tale of Two Cities} (1859); four genres sampled from COCA include blog, news, magazine, and web texts. We only include paragraph-level samples.

Then, we translate the English texts into Chinese using two different LLMs (\textbf{\textit{author}}), i.e., \texttt{Qwen2.5-72B-Instruct}~\citep{qwen25-2025-report} and \texttt{LLama3.3-70B-Instruct}~\citep{Llama3-2024-report}. For each source text and translator LLM, we generate two translations using prompts with two different translation strategies: high-translationese (more literal) and low-translationese (more idiomatic). Hence, a total of 14 paired datasets are created. For each dataset, 1,000 triplets are created for training, 100 for validation, and 100 for testing. The mean translation length of low and high-translationese translations is $86.89 \pm 40.52$ and $83.18 \pm 41.87$, respectively (in tokens, tokenized by \texttt{Qwen2.5}).  Refer to the prompts and examples for two types of translations in the Appendix~\ref{sec:dataset-details}.

% To see whether the unsupervised methods can be utilized in the traditional classification between original and translated texts, we sample 250 original texts and 250 translated texts from LCMC~\citep{mcenery-xiao-2004-lancaster} and ZCTC~\citep{xiao-and-hu-2015-corpus}, and translate them into English. These samples are not strictly paired, but sourced from similar domains, and we expect the detection methods to be able to distinguish the translated texts from the original texts as well.

\subsection{Dataset statistics}
\label{sec:benchmark:dataset-stat}

We compare the statistics of previously studied linguistic features between the low and high-translationese samples, as shown in Table~\ref{tab:benchmark:dataset-stat}.
% , including (1) mean sentence length, (2) mean word length, (3) type-token ratio, (4) frequency of function words, (5) frequency of pronouns, and (6) frequency of punctuations. 
These features are reported to differ between original and translated texts in Chinese. We expect similar differences in our synthetic dataset. 

\begin{table}[ht]
\resizebox{\linewidth}{!}{
    \centering
    \begin{tabular}{llll}
    \toprule
        feature & low & high & $p$-value \\\midrule
        Mean sent. length~$\downarrow$ & 24.701 & 26.465 & 4e-36   \\
        Mean word length~$\downarrow$ & 1.713 & 1.743 & 3e-34   \\
        Type-token ratio~$\downarrow$ & 0.739 & 0.734 & 7e-06   \\
        Freq. of func. words~$\uparrow$ & 0.465 & 0.502 & 4e-193  \\
        Freq. of pron.~$\uparrow$ & 0.052 & 0.059 & 1e-37   \\
        Freq. of punct.~$\downarrow$ & 0.156 & 0.152 & 1e-09  \\\bottomrule
    \end{tabular}}
    \caption{Statistics of the linguistic features for the low- and high-translationese translations. $\downarrow$ indicates that the value of that feature is found to be lower in the translated Chinese compared to non-translated, and $\uparrow$ suggests the opposite.}
    \label{tab:benchmark:dataset-stat}
\end{table}

We conduct independent-sample \textit{t}-tests to test the significance of the difference (see Table~\ref{tab:benchmark:dataset-stat}). 
Out of the 6 features, type-token ratio, function words, pronouns, and punctuations are in alignment with what has been reported in previous literature on Chinese~\citep{xiao-and-hu-2015-corpus,hu-etal-2018-detecting,hu-and-kübler-2021-translated}, suggesting that divergent linguistic patterns between our synthetic low- and high-translationese share similarity with that between the original and translated texts as previously reported and can be used as the starting point for studying T-index.

% The results show that sentences and words in low-translationese translations are longer than those in high-translationese translations. But these statistics are different from what is reported in~\citep{hu-and-kübler-2021-translated} in that translated Chinese text has shorter words and sentences than the original texts. However, for type-token ratio, function words, pronouns, and punctuations, the trend between low and high-translationese translations is consistent with that between original and translated texts, where translationese (at least translated from Indo-European languages) has a lower type-token ratio, uses more function words, more pronouns, but fewer punctuations~\citep{xiao-and-hu-2015-corpus,hu-and-kübler-2021-translated}. 

% The corpus-level statistics indicate that the contrast between low and high-translationese translations, to some extent, is in line with the previous findings on the comparison between original and translated texts, and the differences between the two types of translationese are non-trivial.

\subsection{Human annotation}
\label{sec:benchmark:annotation}

We conduct human evaluation for two purposes: (1) to validate whether the two types of translations exhibit adequate human-perceivable divergence, and (2) to explore how to collect graded human judgments on translationese. We experiment with two annotation methods: pointwise and pairwise annotations. Pointwise annotation provides direct continuous ratings of the translationese degree, and the pairwise annotation can provide an indirect graded judgment by comparing two translations.

\paragraph{Pointwise annotation.} We first experiment on pointwise annotation, each annotator is presented with a source text and a translation. The annotator is asked to rate the translation on a 6-point Likert scale ranging from 0-5, where a higher rating indicates more translationese.
We randomly sample 100 translations and their source texts from the synthetic datasets, with equal number of low and high-translationese samples. Three native Chinese speakers who are
% graduated from 
master students in English/Translation performed the annotation.
% Taking the average of the 3 ratings, only $70.6\%$ of agreement is observed between human annotations and the translation generation strategy. We measure the inter-rater agreement by Fleiss' Kappa, the result of which is $0.248$, suggesting a low agreement. 
Low-translationese samples are rated $1.90 \pm 1.38$ on average, much lower than high-translationese samples, which are rated $3.38 \pm 1.42$ ($p<0.001$). % from an independent \textit{t}-test).
% An independent \textit{t}-test suggests that the difference between the two types of translations is significant ($p<0.001$). 
% However, the inter-rater agreement is low with the Fleiss' Kappa equal to 0.248. 

% Similar low agreement is reported in previous literature regarding human evaluation of translationese. Ten annotators, including 5 professional translators, achieve an average of $78.3\%$ accuracy in distinguishing Italian translated texts from original texts~\citep{baroni-and-bernardini-2005-translationese}.~\citet{wein-2023-human} reports that native speakers distinguish the original English and translated English texts only at a random chance and with an even lower inter-rater agreement ($\mathrm{Kappa}=0.0706$). Our results, along with previous findings, indicate that it is difficult to elicit reliable human judgments concerning translationese by pointwise annotation.

% while the divergence between two classes of texts does exist, in line with findings for MGT detection~\citep{brown-et-al-2020-gpt3,clark-etal-2021-thats,uchendu-etal-2021-turingbench-benchmark,liu-etal-2023-argugpt}.

\paragraph{Pairwise annotation.} We then conduct pairwise annotation, where the annotator is presented with a pair of translations, with the source text, and forced (without a tie) to choose the one that exhibits more translationese. 
We sample 50 triplets from the synthetic data, each containing a source text, a low-translationese translation, and a high-translationese translation. The same three annotators are asked to perform the annotation. We take the majority vote of them as the final human judgment. 
The agreement between human judgments and the generation strategy is $92.0\%$, and the inter-rater agreement measured by Fleiss's Kappa is $0.840$, which indicates very high agreement.

Both pointwise and pairwise annotations confirm that the two groups of translations exhibit valid divergence at the two ends of the translationese continuum, which are also easily detectable by English-Chinese bilingual speakers.

\subsection{More language pairs}
\label{sec:benchmark:more-lps}

% \orange{
We also synthesize French and Germany translations on the basis of the same English source texts with \texttt{Qwen2.5-72B-Instruct}, repeating the generation scheme used for English-Chinese pairs, resulting in 14 \texttt{en-de} and 14 \texttt{en-fr} test sets. Note that we have not validated the quality of generated samples as carefully as what we did to \texttt{en-zh} pairs, and can only provide a preliminary exploration of the cross-linguistic generalization of various measures of translationese with the two language pairs.
% }
\section{Classifying synthetic low- and high-translationese}
\label{sec:generalizable}

\begin{table*}[ht]
\centering
\resizebox{\textwidth}{!}{
    \begin{tabular}{l|lll|lll|ll}
    \toprule
        Language pair & \multicolumn{6}{c|}{\texttt{en-zh}$_{id}$} & \texttt{en-de}$_{ood}$ & \texttt{en-fr}$_{ood}$ \\\midrule
        Author \tiny{(LLM as translator)} & \multicolumn{3}{c|}{\texttt{Qwen2.5-72B-INST}$_{id}$} & \multicolumn{3}{c|}{\texttt{LLama3.3-70B-INST}$_{ood}$} & \multicolumn{2}{c}{\texttt{Qwen2.5-72B}$_{id}$} \\\midrule
        Method $\backslash$ Genre & \textit{OT}$_{id}$ & \textit{Novel}$_{ood}$ & \textit{COCA}$_{ood}$ & \textit{OT}$_{id}$ & \textit{Novel}$_{ood}$ & \textit{COCA}$_{ood}$ & \multicolumn{2}{c}{\textit{All}} \\\midrule
        \multicolumn{9}{c}{\textit{supervised baselines}: models are trained with two classes of translations with their labels} \\\midrule
        DPO (Log-likelihood) & $89.2_{~95.8}$ & $86.9_{~94.5}$ & $82.3_{~89.5}$ & $89.2_{~95.5}$ & $86.8_{~94.4}$ & $80.5_{~88.2}$ & $77.9_{~85.6}$ & $82.6_{~89.7}$ \\
        Bradley-Terry RM & $94.7_{~98.4}$ & $87.6_{~94.2}$ & $87.0_{~93.9}$ & $87.3_{~94.3}$ & $89.9_{~95.6}$ & $87.2_{~93.8}$ & $72.9_{~79.8}$ & $76.5_{~84.6}$ \\
        XLM-RoBERTa & $95.0_{~\text{---}}$ & $83.6_{~\text{---}}$ & $82.1_{~\text{---}}$ & $81.8_{~\text{---}}$ & $87.9_{~\text{---}}$ & $68.9_{~\text{---}}$ & $56.8_{~\text{---}}$ & $60.7_{~\text{---}}$ \\
        SVM w/ ling. feats. & $71.0_{~\text{---}}$ & $73.7_{~\text{---}}$ & $65.1_{~\text{---}}$ & $72.5_{~\text{---}}$ & $69.7_{~\text{---}}$ & $63.8_{~\text{---}}$ & $\text{---}_{~\text{---}}$ & $\text{---}_{~\text{---}}$ \\
        \midrule
        \multicolumn{9}{c}{scoring model $\Tilde{\theta}$ is fine-tuned on high-translationese data: $\Tilde{\theta} \approx \min_{\theta} \mathbb{E}_{(x,y,\Tilde{y}) \in \mathcal{D}}\left[-\Tilde{y} \log (f_{\theta}(x))\right]$} \\\midrule
        Log-likelihood & $80.1_{~87.1}$ & $79.2_{~85.9}$ & $79.0_{~86.1}$ & $80.1_{~87.0}$ & $77.6_{~84.8}$ & $76.9_{~84.4}$ & $71.0_{~77.7}$ & $75.3_{~81.3}$ \\
        Entropy & $73.8_{~77.0}$ & $64.8_{~71.0}$ & $69.1_{~73.7}$ & $71.6_{~74.4}$ & $65.4_{~71.5}$ & $71.0_{~77.9}$ & $60.3_{~63.9}$ & $61.1_{~65.4}$ \\
        Fast-DetectGPT & $74.3_{~80.8}$ & $72.1_{~78.9}$ & $73.1_{~80.5}$ & $72.0_{~79.2}$ & $72.0_{~79.7}$ & $68.5_{~74.9}$ & $70.2_{~76.6}$ & $76.0_{~83.6}$ \\
        Maha. Distance & $55.1_{~54.3}$ & $52.4_{~51.0}$ & $50.2_{~47.2}$ & $54.8_{~53.6}$ & $52.7_{~51.3}$ & $50.3_{~47.5}$ & $51.6_{~50.8}$ & $51.0_{~47.7}$ \\
        Relative Maha. Dist. & $78.3_{~86.0}$ & $70.8_{~77.9}$ & $76.0_{~83.3}$ & $72.8_{~76.9}$ & $73.1_{~80.4}$ & $70.8_{~77.4}$ & $50.1_{~47.9}$ & $50.5_{~49.5}$ \\
        Trajectory Volatility & $63.6_{~69.9}$ & $58.5_{~59.2}$ & $52.1_{~49.6}$ & $60.1_{~60.0}$ & $60.1_{~62.0}$ & $53.1_{~52.0}$ & $51.1_{~49.6}$ & $50.7_{~47.9}$ \\
        \midrule
        \multicolumn{9}{c}{scoring model $\theta$ is fine-tuned on low-translationese data: $\theta \approx \min_{\theta} \mathbb{E}_{(x,y,\Tilde{y}) \in \mathcal{D}}\left[-y \log (f_{\theta}(x))\right]$} \\\midrule
        Log-likelihood & $50.0_{~31.5}$ & $50.0_{~28.6}$ & $50.0_{~25.6}$ & $50.0_{~26.9}$ & $50.1_{~31.0}$ & $50.0_{~24.4}$ & $50.0_{~27.4}$ & $50.0_{~24.9}$ \\
        Entropy & $50.3_{~29.5}$ & $50.0_{~32.8}$ & $50.0_{~28.8}$ & $50.0_{~29.8}$ & $50.1_{~32.6}$ & $50.0_{~25.6}$ & $50.0_{~38.3}$ & $50.0_{~37.6}$ \\
        Fast-DetectGPT & $53.5_{~49.0}$ & $50.1_{~39.3}$ & $50.0_{~42.2}$ & $51.1_{~40.9}$ & $50.0_{~41.9}$ & $50.0_{~40.0}$ & $50.0_{~28.8}$ & $50.1_{~22.2}$ \\
        Maha. Distance & $56.3_{~55.0}$ & $54.6_{~54.3}$ & $55.1_{~56.1}$ & $55.8_{~51.5}$ & $55.8_{~55.2}$ & $54.0_{~54.4}$ & $51.3_{~49.6}$ & $53.3_{~53.2}$ \\
        Relative Maha. Dist. & $74.8_{~81.9}$ & $69.1_{~74.9}$ & $72.2_{~78.5}$ & $71.3_{~74.5}$ & $70.0_{~77.0}$ & $66.6_{~72.3}$ & $50.2_{~48.9}$ & $50.7_{~49.9}$ \\
        Trajectory Volatility & $58.8_{~59.6}$ & $57.3_{~58.2}$ & $61.8_{~65.3}$ & $57.0_{~56.6}$ & $57.7_{~58.7}$ & $60.0_{~62.4}$ & $51.8_{~50.8}$ & $51.7_{~50.8}$ \\
        \midrule
        \multicolumn{9}{c}{likelihood ratios of $\Tilde{\theta}$ and $\theta$: $\log \mathrm{P}_{\Tilde{\theta}}(y \mid x) - \log \mathrm{P}_{\theta}(y \mid x)$} \\\midrule
        \textbf{Translationese Index} & $95.8_{~99.2}$ & $92.7_{~97.8}$ & $95.2_{~98.8}$ & $93.0_{~97.9}$ & $93.5_{~98.3}$ & $90.5_{~96.2}$ & $79.0_{~85.5}$ & $82.3_{~90.0}$ \\
        \bottomrule
    \end{tabular}
}
\caption{Results (accuracy reported with auroc as subscript) for the binary classification between low-translationese and high-translationese. All scoring and classifying models are trained on source-target pairs of \textit{Oliver Twist} translated by \texttt{Qwen2.5-72B-Instruct}. \textit{id} indicates the in-domain test set, and all other columns denote cross-domain test sets (all results are the average across 3 random seeds).}
    \label{tab:generalizable:results}
\end{table*}

% Since the two classes of translations in the toy benchmark are validated to be divergent by both annotation methods, 
We start with a binary classification task, distinguishing translations of low-translationese  from those with high-translationese. Specifically, we evaluate T-index along with several unsupervised (\S\ref{sec:generalizable:unsupervised}) and supervised (\S\ref{sec:generalizable:supervised}) baselines in the cross-domain settings.

% A qualified translationese measurement method should almost perfectly discriminate the two classes even in unseen domains.

% To investigate how well T-index can measure translationese, we start with a binary classification task, distinguishing high-translationese samples from low-translationese ones. Furthermore, we want to compare the cross-domain generalization of various supervised and unsupervised methods with T-index.

\subsection{Unsupervised baselines}
\label{sec:generalizable:unsupervised}

We first fine-tune an LLM on translations from a specific domain. Then we can use this model as a proxy for its training distribution, and use a scoring function relying on features given by the scoring model to estimate the resemblance between the test sample and the training distribution. Let's say a scoring model is fine-tuned on high-translationese data, then this model is likely to assign higher log-likelihood to high-translationese then low-translationese.

% Following the setups in OOD detection, we first SFT a pre-trained LLM as the scoring model. With the training distribution, the fine-tuned model is expected to distinguish the ID and OOD samples using certain scoring functions. Analogously, we can SFT an LLM on the low-translationese samples, and this model might be able to distinguish the low-translationese translations (ID) from the high-translationese ones (OOD).

\paragraph{Scoring models.}
% To evaluate cross-domain generalization, we fix the sources sampled from \textit{Oliver Twist} and translations generated by \texttt{Qwen2.5-72B-Instruct} as the only training domain. 
On paired samples from the training domain, we SFT \texttt{Qwen2.5-0.5B} with the translation task to obtain two scoring models. One is fine-tuned on low-translationese translations, and the other on high-translationese translations.
% with the same set of hyperparameters. 
% The scoring models are fine-tuned with a global batch size of 32 for 800 steps at a learning rate of 1e-06. Fine-tuning only takes around five minutes on four 3090 GPUs.
% We refer to the two models as ``low-translationese model'' and ``high-translationese model''.

\paragraph{Scoring functions.}
We include several scoring functions that are tested to be useful in machine-generated text and out-of-distribution detection as unsupervised baselines and re-implement them for translationese measurement, 
% Here, we briefly describe the intuition of each algorithm when the scoring model is a high-translationese model.
including three logits-based functions: Log-likelihood, Entropy, and Fast-DetectGPT~\citep{bao-etal-2024-fast}, Mahalanobis Distance~\citep{ren-etal-2023-outofdistribution}, and three embedding-based functions: Relative Mahalanobis Distance~\citep{ren-etal-2023-outofdistribution}, and Trajectory Volatility~\citep{wang-etal-2024-embedding} (see Appendix~\ref{sec:further-discussion-binary} for details).

\subsection{Supervised baselines}
\label{sec:generalizable:supervised}

The methods above do not rely on the labels of the training samples. We also include supervised baselines trained explicitly with labeled samples, including DPO-aligned Reward~\citep{rafailov-etal-2023-direct,rafailov-etal-2024-dpo-reward}, Bradley-Terry Reward Model~\citep{bradley-terry-1952-rank,ouyang-etal-2022-training}, XLM-RoBERTa~\citep{conneau-etal-2020-unsupervised}, and SVM with linguistic features~\citep{hu-and-kübler-2021-translated}.

% Along with methods drawn from MGT and OOD detection, we also include \texttt{XLM-RoBERTa}~\citep{conneau-etal-2020-unsupervised} and SVM with linguistic features~\citep{hu-and-kübler-2021-translated} as supervised baselines. XLM-R and SVM will be trained supervisedly on the paired low and high-translationese samples.

\subsection{Evaluation metrics} 
We report \textbf{\textit{accuracy}} as our metric for the binary classification task, where the majority baseline is $50\%$. For methods that yield a continuous score, we compute the threshold that maximizes the accuracy on each test set. We also report \textbf{\textit{auroc}}, ranging from 0 to 1, where 1 indicates perfect discrimination. For binary classifiers, we only report accuracy.

\subsection{Results}
\label{sec:generalizable:results}

We present the results of various measuring (or detecting) methods in Table~\ref{tab:generalizable:results}. We mainly evaluate the cross-domain generalization, where models are trained on the single domain of \textit{Oliver Twist} (OT) translated by \texttt{Qwen2.5-72B-Instruct}
% \orange{
from English to Chinese, which is denoted as \textit{id} (in-domain). Any other test set from a different genre, translated by another LLM, or from a different language pair, is denoted as \textit{ood} (out-of-domain).
% }

\paragraph{T-index is generalizable.}
On the in-domain test set, T-index almost perfectly classifies low-translationese and high-translationese. Three supervised methods, DPO with log-likelihood, Bradley-Terry RM and XLM-R, can also achieve an accuracy around $90\%$.
When it comes to the cross-domain test sets, an increasing gap occurs between T-index and the supervised baselines. 
T-index remains highly discriminative under the influence of genre shift, while a drop in accuracy can be observed in supervised baselines as the genre shift away from the training domain. 
The limited generalizability of supervised measures is possibly due to learning domain-specific features rather than translationese features~\citep{amponsah-kaakyire-etal-2022-explaining}. 
The same observation applies to the author shift. Yet in this setting, T-index also undergoes a decrease in accuracy. 

% \orange{
When evaluated on English-Germany/French translations, the performance drops notably for even T-index. But transfer to these language pairs is non-trivial for DPO, RM, and T-index, which might result from the multilinguality of the pre-trained base model.
We leave detailed cross-lingual analysis to future work.
% }
% XLM-R might learn domain-specific features rather than translationese features~\citep{amponsah-kaakyire-etal-2022-explaining}, so that it fails to generalize to other domains.
% However, T-index is less robust to the translations produced by another LLM with a 5-point gap. However, other baselines even underperform in the \textit{i.d.} test set.

\paragraph{Logits-based features are more discriminative.}
Performance varies for unsupervised methods depending on the features and models used for scoring. When $\Tilde{\theta}$, fine-tuned on high-translationese, is used for scoring, logits-based methods generally outperform embedding-based methods.
The simple log-likelihood can already achieve around 80\% on most \texttt{en-zh} translations, and the other two methods also perform above random guessing. 
We suspect that embeddings might encode the semantics of translations, instead of lexical choice and word order that shape translationese, which can be better captured by the logits of LMs.

% can discriminate the two classes at $\sim$$80\%$ of accuracy and are insensitive to the distribution shift in the test data, which is better than embedding-based methods. We suspect that the lexical choice and word order that shape translationese might be blurred in the average of embeddings.

% Notably, embedding-based methods are reported to surpass logits-based methods in OOD detection~\citep{ren-etal-2023-outofdistribution}, when the ID samples are from WMT15~\citep{bojar-etal-2015-WMT} and OOD samples are from MTNT~\citep{michel-neubig-2018-mtnt}. WMT15 contains more formal texts, while MTNT consists of web-based ``noisy'' texts. However, in our setting, translationese is shaped by the lexical choice and word order, which might be blurred in the average neural representations.

\paragraph{Logits might encode stylistic features in the training distribution.}
If the scoring model is $\theta$ fine-tuned with low-translationese data, the logits-based functions even yield scores negatively correlated with the degree of translationese (with an AUC under 50). We attribute this observation to the fact that the model fits better to the high-translationese fine-tuning data, with a lower training loss, and $\theta$ fine-tuned on low-translationese still assigns higher probability to the high-translationese translations.

The log-likelihood of an LM somehow encodes the stylistic feature that fits its training distribution\footnote{The training distribution can be pre-training or post-training distribution}, making it sensitive to the scoring models. Therefore, T-index can tease apart the stylistic shift caused by confounding factors, by ensembling the likelihoods of $\theta$ and $\Tilde{\theta}$ that share the same distribution regarding genre and author (see Appendix~\ref{sec:empirical-confirm}).

\section{Measuring translationese in the wild}
\label{sec:graded}

The synthetic translations used in the previous section are elicited with specifically designed prompts, under-representing real-world MTs, where the textual differences might be more nuanced. In this section, we use translations in the wild to investigate whether the continuous scores of T-index can serve as a graded measurement of translationese that aligns with human judgments.

% (\S\ref{sec:t-index-human}). Finally, we explore the correlation between T-index and existing MT QE metrics (\S\ref{sec:t-index-qe}).

% \subsection{T-index aligns with human annotations}
% \label{sec:t-index-human}

\subsection{Collection of human annotations}

We sample 50 texts from the same source texts in the 
% synthetic dataset
previous section
and translate them into Chinese with 3 MT systems or LLMs (with the vanilla prompt) randomly selected from a pool of seven MT systems/LLMs (see Appendix~\ref{sec:dataset-details}), obtaining 3 translations per source text. 
Note that we name these translations ``in the wild" since we do not instruct the MT-systems or LLMs to produce translations of specific levels of translationese. 
Thus, all translations contain an arbitrary level of translationese that we need to find out, either with human annotators or T-index.

% For each source text, we can have 3 pointwise ratings and 3 pairwise rankings.
Following the annotation schemes in the synthetic dataset, we collect pointwise human ratings and pairwise judgments for in-the-wild MTs:
\begin{itemize}[leftmargin=*]
    \setlength{\itemsep}{0pt}
    \item \textbf{Pointwise}: we ask over 30 master students in Translation Studies from prestigious Chinese universities to rate the degree of translationese from 0-5 on 150 (\textit{source}, \textit{translation}) pairs. Each text is rated by more than 10 raters, and then we use the mean rating of each text as the result.
    \item \textbf{Pairwise}: a different set of five annotators is asked to choose the high-translationese translation, given 150 (\textit{source}, \textit{translationA}, \textit{translationB}) triples.  We take the majority vote as the ground truth.
\end{itemize}

% Annotators are also asked to extract spans containing the most translationese, which can be used 
% % we will explore in future work 
% for token/span-level translationese measurement. 
Each annotation, for both pointwise and pairwise, takes 1-2 mins. For every 50 annotations, the annotator is compensated with 60 Chinese \textit{yuan}.
% \orange{During annotation, annotators are also asked to extract spans containing the most translationese. 
% For one reason, we use this method to make annotators more focused on the task; for the other reason, these spans can be used for a finer-grained measurement, which we will explore in future work
% }

% We only sample 1 translation for each source. 

\begin{figure}[ht]
    \centering
    \includegraphics[width=\linewidth]{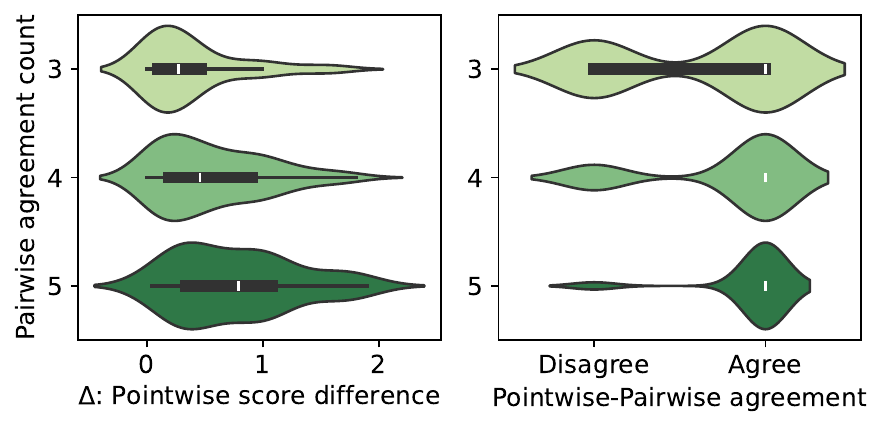}
    \caption{Distribution of human annotation agreement patterns. Left: pointwise score differences by pairwise agreement level. Right: pairwise agreement counts by pointwise-pairwise consistency. Higher annotator consensus corresponds to larger rating differences between translation pairs.}
    \label{fig:graded:human-anno}
\end{figure}

% \orange{
In pairwise annotation, the inter-rater agreement (Fleiss' Kappa = 0.287) is substantially lower than that observed in the synthetic dataset (0.840).
Upon observing the annotations closely, we interpret the lower Kappa as the following: for some documents,  translation A may demonstrate more translationese in certain positions, while translation B may show translationese in other places, thus making it difficult to make a decision at the document level. 
However, for other documents, the choice might be easier since one translation clearly demonstrate more translationese as a whole.  
% }

% \orange{
% This indicates that human annotators face greater uncertainty when identifying translationese in real-world translations. The disagreement and uncertainty in annotations highlight the graded nature of translationese, suggesting it exists on a spectrum rather than as a binary phenomenon.
% } \blue{HH: disagreement and uncertainty are different things.}

% \blue{
Therefore, we categorize pairwise judgments into three groups based on the number of agreement for a given triple: 
five (unanimous agreement), four, and three out of the five annotations. 
% }
% inter-annotator agreement (pairwise agreement count), as shown in Table~\ref{tab:graded:results}: 
% (1) unanimous agreement among all five annotators; (2) agreement among four annotators; and (3) agreement among three annotators. 
This categorization captures the difficulty and disagreement in pairwise annotations, which we then compare with pointwise ratings (see Figure~\ref{fig:graded:human-anno}).

% \orange{
We observe consistent trends between pairwise and pointwise annotations. As pairwise annotation certainty increases (higher agreement among annotators), the differences in pointwise ratings between the chosen and rejected translations become more pronounced. Similarly, agreement between the two annotation methods increases when the paired translations exhibit greater differences in translationese characteristics.
We believe that different levels of disagreement among annotators (manifest in the pairwise agreement count) demonstrate that different shades of translationese are indeed observed by human annotators.
% }

\subsection{T-index aligns with human judgments}

For pairwise evaluation, we compute T-index for each translation and compare the values. The sample in a pair with a greater value is considered to be the prediction given by T-index. 
We further include LLM-as-a-judge method for comparison: 
% several baselines that are often used in LLM evaluation, including two off-the-shelf general reward models: \texttt{InternLM2-20B-Reward}~\citep{cai-etal-2024-internlm2} and \texttt{Skywork-Reward-Gemma2-27B-v0.2}~\citep{liu-etal-2024-skywork}; and LLM-as-a-judge methods: 
\texttt{LLama3.3-Instruct-70B}~\citep{Llama3-2024-report} and \texttt{Qwen2.5-Instruct-72B}~\citep{qwen25-2025-report} (see prompts in Appendix~\ref{sec:anno-judge}).

% \paragraph{Scoring models and baselines.} 
In \S\ref{sec:generalizable}, we only use data from one single domain of the synthetic dataset for SFT to test the generalization. In this section, we compare scoring models trained with different data: (1) Unpaired samples: SFT data for the two models come from two different domains; (2) Single domain: SFT data for the two models are paired, but only one domain of data is used; (3) Mixed domain: We mix pairs from all domains together with the sample size ranging from 1k to 5k.
We also fine-tune BT RM and DPO-aligned models on the same 5k samples as T-index for comparison.

% \paragraph{Results.} The translationese index is able to predict the human judgment with a high accuracy of $89.7\%$ when all 5 annotators agree (see Table~\ref{tab:llr-pairwise}). The agreement drops as the human variance occurs, while the $\Delta$ translationese index still predicts the human uncertainty (see Figure~\ref{fig:agreement-analysis}). Notably, the general reward models are not reliable in predicting the translationese, and the LLM-as-a-judge can achieve around $80\%$ accuracy when all 5 annotators agree but relying much more parameters. 

% \paragraph{Agreement with pairwise annotations.} 

\begin{table}[ht]
    \centering
    \resizebox{\linewidth}{!}{
    \begin{tabular}{lccc|c}
    \toprule
        Pairw. agreement count & 3 & 4 & 5 & Pointw.\\\midrule
        \# samples & N=45 & N=66 & N=39 & N=150 \\\midrule
        & \multicolumn{3}{c|}{Agreement$\uparrow$} & Pearson's \textit{r}$\uparrow$\\\midrule
        T-index & & & \\
        ~~~~w/ unpaired (1k) & 64.4 & 66.7 & 82.1 & 34.8 \\
        ~~~~w/ single-dom. (1k) & 68.9 & 74.2 & \textbf{84.6} & 34.6 \\
        ~~~~w/ mixed-dom. (1k) & 68.9 & \textbf{77.3} & 82.1 & 32.0 \\
        ~~~~w/ mixed-dom. (3k) & 55.6 & 74.2 & 74.4 & 39.2 \\
        ~~~~w/ mixed-dom. (5k) & 57.8 & 74.2 & \textbf{84.6} & \textbf{41.8} \\\midrule
        BT RM (5k) & 57.8 & 72.7 & 76.9 & 40.7 \\
        DPO-aligned (5k) & 62.2 & 66.7 & 76.9 & 19.7 \\\midrule
        % \texttt{InternLM2-Reward} & 60.0 & 65.1 & 64.1 & --- \\
        % \texttt{Skywork-RM-Gemma2} & 57.7 & 65.1 & 58.9 & --- \\
        \texttt{LLama3.3-Instruct} & 57.7 & 53.0 & 79.4 & --- \\
        \texttt{Qwen2.5-Instruct} & 62.2 & 68.1 & \textbf{84.6} & --- \\\midrule
        Human Pointwise & 60.0 & 77.3 & 92.3 & --- \\
    \bottomrule
    \end{tabular}}
    \caption{Agreement between automated methods with majority votes in the pairwise annotation (agreement reported) and correlation evaluated against mean ratings in pairwise annotation (Pearson's $r$ reported).}
    \label{tab:graded:results}
\end{table}

\paragraph{Most automatic methods can predict human pairwise judgments.}
The results in Table~\ref{tab:graded:results} demonstrate that most automated methods achieve above-chance performance in agreement with human pairwise judgments. Among the evaluated methods, T-index and \texttt{Qwen2.5-Instruct-72B} achieve the highest accuracy of 84.6\% on pairs with unanimous agreement among all five human annotators. However, a performance gap of approximately 8 percentage points remains compared to the agreement between human pairwise judgments and human pointwise ratings.

\paragraph{T-index correlates moderately with human pointwise ratings.}
When evaluated against continuous human pointwise ratings, we measure correlation strength using Pearson's $r$. We find that T-index and BT RM achieve moderate-to-high correlations ($\sim$0.4) with human mean ratings. 
The results suggest that training on paired translationese data, as showcased by the synthetic datasets, can help the models capture the gradience of translationese.
Notably, higher accuracy on pairwise judgments does not necessarily translate to stronger correlation with pointwise ratings. For example, T-index trained on 1k samples from a single domain achieves similar pairwise accuracy to T-index trained on 5k samples, yet lags 8 percentage points behind in pointwise correlation.

\paragraph{T-index is data-efficient.}
Ablation experiments on training data in the upper half of Table~\ref{tab:graded:results} demonstrate that T-index is robust across different data conditions. Even when scoring models are trained on only 1k samples from different domains (unpaired), T-index can still effectively predict the degree of translationese. While increasing the amount of training data improves correlation with human pointwise ratings, it does not notably impact pairwise agreement performance.

\subsection{T-index complements existing automatic QE metrics}
\label{sec:graded:qe}

We further explore the correlation between T-index and existing automatic QE metrics, including 3 reference-based metrics, xCOMET~\citep{guerreiro-etal-2024-xcomet}, BLEURT-20~\citep{pu-etal-2021-learning}, and BLEU~\citep{papineni-etal-2002-bleu}, and one reference-free metric, COMET-Kiwi22~\citep{rei-etal-2022-cometkiwi}.

\begin{table}[ht]
    \centering
    \resizebox{\linewidth}{!}{
    \begin{tabular}{lll}
    \toprule
        Dataset & Source (en) & Translation (zh) \\\midrule
        Standard & original English & HT Chinese \\
         & \multicolumn{2}{l}{\textit{\footnotesize source is original English, and translation is HT reference.}} \\
        Reverse & MT English & original Chinese \\
         & \multicolumn{2}{l}{\textit{\footnotesize translation is original Chinese, and source is MT from Chinese.}} \\
        Back-translate & MT English & HT Chinese \\
         & \multicolumn{2}{l}{\textit{\footnotesize translation is translated Chinese, and source is back-translation.}} \\
    \bottomrule
    \end{tabular}}
    \caption{Three conditions used for the overall MT QE task. The standard MT evaluation is sampled from Flores101~\citep{goyal-etal-2022-flores}; the original Chinese in the reverse test set is from LCMC~\citep{mcenery-xiao-2004-lancaster}; the HT Chinese in the back-translation test set is from ZCTC~\citep{xiao-and-hu-2015-corpus}. The MT English is generated by Google Translate.}
    \label{tab:graded:qe-data}
\end{table}

\begin{figure*}[ht]
    \centering
    \includegraphics[width=\linewidth]{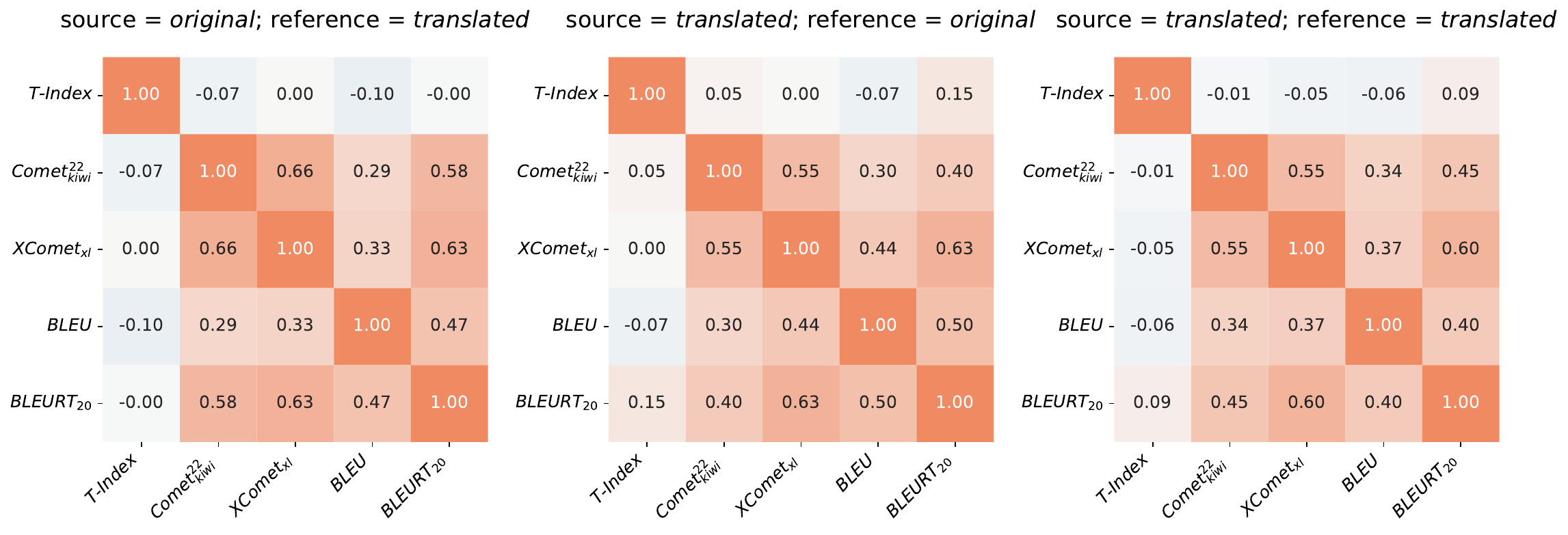}
    \caption{Correlation between the T-index and automatic MT QE metrics. To mitigate the bias that translationese in references or sources leads to, we also use the reverse test set and the back-translation test set as control groups.}
    \label{fig:graded:overall-qe}
\end{figure*}

The correlation is computed under 3 conditions (see Table~\ref{tab:graded:qe-data}): (1) standard MT evaluation, where the source is originally written and the reference is human translation; (2) reverse test set, where zh-en translations are used for en-zh evaluation, so there is no translationese in references; and (3) back-translation test set, where the source is back-translated from the human translated references. For each condition, we sample 1000 source-reference pairs, and we use 5 LLMs to generate MTs (see Appendix~\ref{sec:dataset-details}).

Results in Figure~\ref{fig:graded:overall-qe} show only a weak correlation between T-index and existing automatic QE metrics, which indicates that translationese features are not yet covered by them. T-index can therefore be used as a complementary metric for MT QE.

\section{Related Work}
\label{sec:related}

\paragraph{Translationese identification.}
Linguistic theories hypothesize that certain features could serve as indicators of translationese~\citep{toury-1995-descriptive,baker-1996-corpus}. These features are operationalized through feature engineering in the binary classification task between original and translated texts. For instance,~\citet{volansky-et-al-2015-features} and~\citet{hu-and-kübler-2021-translated} train machine-learning classifiers with linguistic features, such as type-token ratio, POS \textit{n}-grams, and grammar rules.  Some information-theoretic features~\citep{lembersky-etal-2012-language,rubino-2016-information,bizzoni-etal-2020-human} are also used to identify the translationese. Therefore, these features are often used to estimate the translationese level of a text. For example, \citet{li-etal-2025-literalism} uses perplexity, lexical density, and length variance for estimate translationese to filter high-quality training data. However, these methods mostly provide corpus-level statistics, not applicable to MT QE which often requires sample-level information.

Mostly related to our work, \citet{freitag-etal-2022-natural} contrast the likelihood of a natural LM and translationese LM to estimate the naturalness of the MT training data, but only validate it on the classification between original and translated texts without discussing its potential to be used as a finer-grained measurement.

\paragraph{Translationese and translation quality.}
Though nonnative, translationese it is not necessarily a defect in translation. In translation theory, translators can even purposely foreignize the translations relating to the style and culture in the source language~\citep{venuti-1994-translators}. For HT, translationese is not an obvious indicator of poor translation quality~\citep{kunilovskaya-lapshinova-koltunski-2019-translationese}. 
The same goes for MT, the mild translationese is also acceptable for better faithfulness and accuracy~\citep{freitag-etal-2022-natural,flamich-etal-2025-cannot}. However, the rigid translationese, more frequently observed in MTs~\citep{freitag-etal-2019-ape,bizzoni-etal-2020-human,luo-etal-2024-diverge}, is what should be penalized. Translationese is part of the overall translation quality, but it is just one of the many factors that affect the quality. 
% However, in this paper, we isolate the translationese from other factors.

\paragraph{Unsupervised methods in text classification.}
Though machine-generated text (MGT) and out-of-distribution (OOD) detection are two text classification tasks, the classification often utilizes unsupervised methods to score two classes of texts. Scores rely on the internal features of the scoring models. Models can be seen as the proxies of training distributions, and the scores quantify how the test samples resemble the training distribution, which is aligned with the objective of translationese measurement.
For MGT detection, \citet{gehrmann-etal-2019-gltr} and \citet{solaiman-etal-2019-release} pioneer logits-based features, such as probability, to distinguish between human-written and machine-generated texts. \citet{mitchell-etal-2023-detectgpt} and \citet{bao-etal-2024-fast} further propose perturbed-based methods. For OOD detection, the detection relies on signals, such as different levels of confidence, usually estimated by probabilities and entropy~\citep{hendrycks-and-gimpel-2017-baseline,ren-etal-2019-likelihood,hendrycks-etal-2020-pretrained,arora-etal-2021-types} or different geometric properties of hidden states, quantified by distance or between-layer changes~\citep{ren-etal-2023-outofdistribution,jelenic-etal-2024-outofdistribution,wang-etal-2024-embedding}. These methods are potentially applicable to translationese detection as well.

\section{Conclusion and future work}
\label{sec:conclusion}

In this paper, we aim to develop a graded and generalizable measure of translationese. To this end, we reframe translationese measurement as a comparative task between different translations of the same source text, rather than binary classification between translated and non-translated text. Under this new formulation, among evaluated methods, T-index (likelihood ratios of two contrastively fine-tuned LLMs) has the best generalizability and alignment with human ratings and judgments. 

We also show that T-index is weakly correlated with several automatic MT QE metrics, suggesting that T-index can be a complementary measure to existing MT QE metrics, which is especially important when existing MT QE metrics focus more on the accuracy and become less reliable in evaluating MTs of higher quality produced by LLMs~\citep{agrawal-etal-2024-automatic-metrics,kocmi-etal-2024-findings}.

Our work complements previous studies that view translationese as features distinguishing translated from non-translated texts by a new perspective: translationese can also relate to readers' linguistic intuition directly. Building on this, future work could investigate more scalable annotation methods to capture this intuition through comprehensive human experiments. These annotations would then enable the development of finer-grained automated measures for MT system evaluation and post-training.

Beyond MT, this work can be extended to other natural language generation tasks. While binarized features like accuracy or factualness can be more easily automated, there remains a class of features that require graded measurement. Features like translationese or naturalness are more nuanced yet equally essential to the reading experience of LLM-generated texts, opening up important directions for future automated evaluation methods.

\section*{Limitations}

We primarily verified T-index on the English-Chinese language pair. Further research is needed to see whether the results are generalizable to other language pairs, especially when the two languages are similar and translationese is more difficult to define and detect.
% Also, we did not provide a theoretical proof of the T-index and its broader applicability. 
We only conduct preliminary human experiments, collecting a limited number of human annotations both in the synthetic benchmark and the in-the-wild dataset.
Future research can build upon our work to collect more human annotations to examine human (dis)agreement in greater depth.
% We haven't explored the human disagreement in translationese measurement in depth.

\section*{Ethics Statements}
The source texts used in this paper are sampled from classic novels and well-curated corpora, and the translations are produced by open-sourced LLMs. Therefore, we believe that there is no risk of leakage of personal identifiable information or any ethical issues. The data used in this paper are only intended for research concerning the MT evaluation and should not be interpreted otherwise.

\section*{Acknowledgements}
We thank our annotators for their help in data annotation. 
This work is funded by Shanghai Pujiang Program (22PJC063) awarded to Hai
Hu and the General Program of National Natural Science Foundation of 
China (62176153) awarded to Rui Wang.

\bibliography{custom}

\appendix

\section{Empirical confirmation of T-index}
\label{sec:empirical-confirm}

\begin{figure*}[ht]
    \centering
    \includegraphics[width=\textwidth]{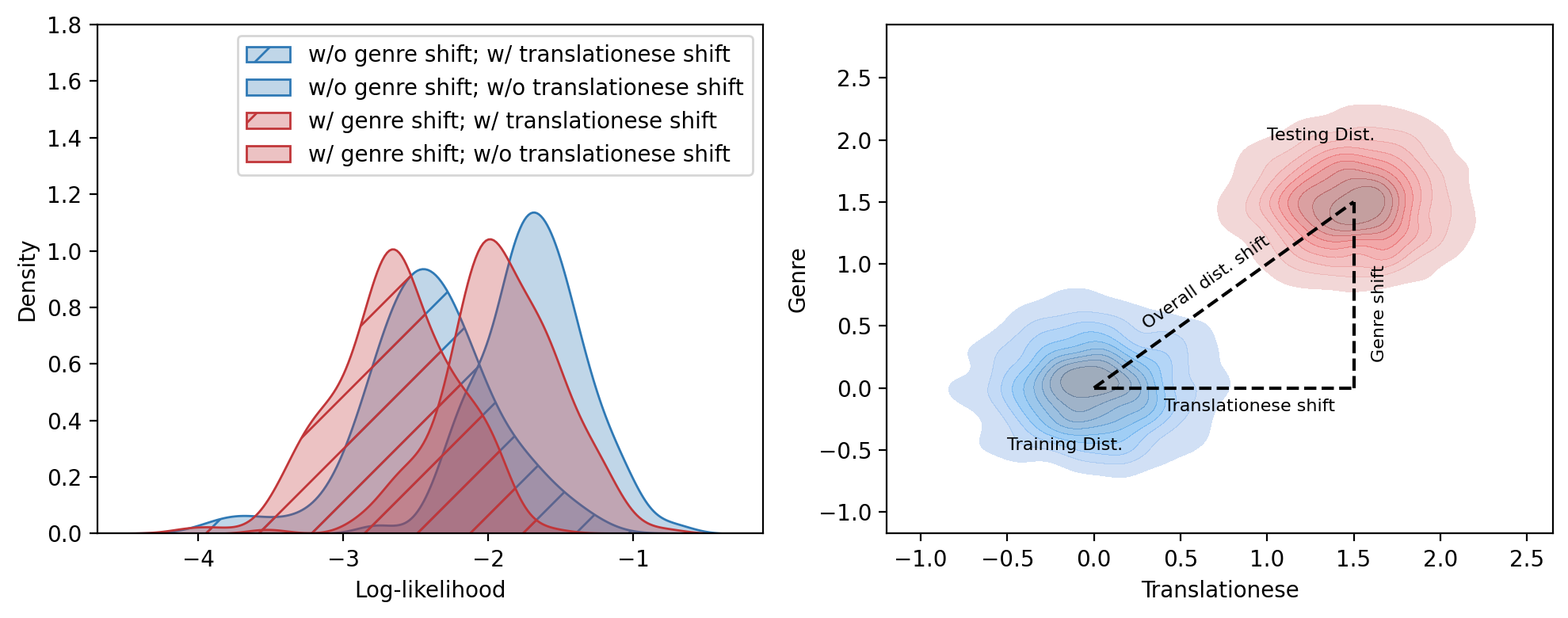}
    \caption{In translationese measurement, the feature wanted is translationese. However, other types of distribution shift might be encoded in log-likelihood as well,  illustrated by the left-side figure. The model assigns the highest probabilities to samples from the same distribution as the training data, and assigns lower probabilities when testing samples shift caused by both translationese and genre. The right-side figure intuitively illustrates that the overall distribution shift can be decomposed into independent components, which is the fundamental assumption of LLR.}
    \label{fig:appendix:llr-intuition}
\end{figure*}

To confirm that different components are independent, we quantify the three types of shifts between the training distribution and the testing distribution. First, we use the mean log-likelihood (MLL) as the statistics to measure a distribution represented by the dataset $\mathcal{D}$, which is defined as:

\begin{align}
    \mathrm{MLL}(\mathcal{D}; \theta) = \frac{1}{|\mathcal{D}|} \sum_{(x,y) \in \mathcal{D}} \frac{1}{|y|} \log \mathrm{P}_{\theta}(y \mid x).
\end{align}

Then, we obtain the MLL of the training distribution as a reference value, where $\texttt{ref\_MLL} = \mathrm{MLL}(D_{g_i,a_i,t_i}; \theta_{g_i,a_i,t_i})$. With this reference, we can measure the distribution shift between $D_{g_i,a_i,t_i}$ and testing distribution $D_{g_j,a_j,t_j}$ by the difference between $\mathrm{MLL}(D_{g_j,a_j,t_j; \theta_{g_i,a_i,t_i}})$ and \texttt{ref\_MLL}.

To measure the shift of each independent component, we keep the other two components the same as the training distribution but change the targeted one to the value of the testing distribution. Then the four types of shifts are defined as follows\footnote{Note that \texttt{o\_shift} is the overall shift, \texttt{g\_shift} is the genre shift, \texttt{a\_shift} is the author shift, and \texttt{t\_shift} is the translationese shift, and the scoring model here is trained on $D_{g_i,a_i,t_i}$, for simplicity, we omit the parameter $\theta_{g_i,a_i,t_i}$ in the notation.}:
\begin{align}
    \texttt{o\_shift} = \mathrm{MLL}(D_{g_j,a_j,t_j}) - \texttt{ref\_MLL} \notag \\
    \texttt{g\_shift} = \mathrm{MLL}(D_{g_j,a_i,t_i}) - \texttt{ref\_MLL} \notag \\
    \texttt{a\_shift} = \mathrm{MLL}(D_{g_i,a_j,t_i}) - \texttt{ref\_MLL} \notag \\
    \texttt{t\_shift} = \mathrm{MLL}(D_{g_i,a_i,t_j}) - \texttt{ref\_MLL} \notag
\end{align}

We obtain 28 models fine-tuned on the 28 datasets and evaluate them on all 28 tests, resulting in 784 observations in total. We run an OLS regression on the four types of shifts, where $\texttt{overall\_shift} \sim \texttt{genre\_shift} + \texttt{author\_shift} + \texttt{translationese\_shift}$. The linear model explains 97.3\% of the variance in the \texttt{overall\_shift}; \texttt{genre\_shift} ($\beta = 1.0129$, $p < 0.001$), \texttt{translationese\_shift} ($\beta = 0.9813$, $p < 0.001$), and \texttt{author\_shift} ($\beta = 0.7527$, $p < 0.001$) are all significant predictors. The variance inflation factor (VIF) of the three predictors is $1.001$, $1.000$, and $1.002$, respectively, indicating that the three predictors are independent. The assumption (a) of LLR is empirically confirmed.

With the independence of the three types of shifts, the follow-up question is whether a contrastively fine-tuned model can cancel out the genre and author shifts, the assumption (b). First, we define a model $\theta_{g_i,a_i,t}$ as the model fine-tuned on the dataset $\mathcal{D}_{g_i,a_i,t}$. The contrastively fine-tuned model is $\theta_{g_i,a_i,\Tilde{t}}$. We run a paired \textit{t}-test for the values of \texttt{genre\_shift} and \texttt{author\_shift} given by $\theta_t$ and $\theta_{\Tilde{t}}$, which turns out to be insignificant for \texttt{genre\_shift} ($p = 0.808$) but significant for \texttt{author\_shift} ($p < 0.001$). The results also explain the performance drop of LLR with the author shift in Table~\ref{tab:generalizable:results}.

The empirical confirmation suggests that for each model, the log-likelihood can be decomposed into three independent components. However, the unwanted shifts are mostly, but not completely, canceled out by a contrastively fine-tuned model. 

\section{Details of the datasets}
\label{sec:dataset-details}
\subsection{Synthetic benchmark}\label{sec:details:synthetic:benchmark}
Please refer to Table~\ref{tab:prompts} for the prompts used to generate low- and high-translationese in the synthetic benchmark and translation examples.

\begin{CJK*}{UTF8}{gbsn}
\begin{table*}[htbp]
\centering
\begin{tabular}{p{0.15\linewidth}p{0.7\linewidth}}
\toprule
\textbf{Type} & \textbf{Content} \\
\midrule
\multicolumn{2}{l}{\textbf{Translation Prompts}} \\
\midrule
Low-translationese & 请把以下文本翻译为中文。译文必需符合中文表达，多用小句、流水句以及中文俗语，不一定非要忠实于原文。请直接返回译文。 \\
& \textit{[Please translate the following text into Chinese. The translation should be idiomatic Chinese, favoring shorter sentences, run-on sentences, and Chinese colloquialisms. It doesn't have to be strictly faithful to the source text. Please return the translation directly.]} \\
\midrule
High-translationese & 请把以下文本翻译为中文。译文必需忠实于原文，不要为了中文用语习惯对原文做任何修改。请直接返回译文。 \\
& \textit{[Please translate the following text into Chinese. The translation must be faithful to the original text, and no modifications should be made to the text to fit conventions in the Chinese language. Please return the translation directly.]} \\
\midrule
\multicolumn{2}{l}{\textbf{Translation Examples}} \\
\midrule
Source text & The three spectators seemed quite stupefied. They offered no interference, and the boy and man rolled on the ground together; the former, heedless of the blows that showered upon him, wrenching his hands tighter and tighter in the garments about the murderer's breast, and never ceasing to call for help with all his might. \\
\midrule
Low-translationese & 三个旁观者愣住了，他们没敢上前阻拦，只见那男孩和男人在地上滚成一团。男孩不顾雨点般落在身上的拳打脚踢，死死揪住凶手胸前的衣服，一边拼命地高声呼救，一边越揪越紧。 \\
\midrule
High-translationese & 那三个观众似乎相当惊呆了。他们没有进行干涉，那个男孩和男人在地上滚作一团；前者不顾落在他身上的拳打脚踢，越发力气地抓紧凶手胸前的衣裳，并且一直用尽全力呼救。 \\
\bottomrule
\end{tabular}
\caption{Translation prompts and examples for low- and high-translationese in the synthetic dataset.}
\label{tab:prompts}
\end{table*}
\end{CJK*}

\subsection{Datasets in-the-wild}
For the in-the-wild MTs for human annotation, we choose 7 systems of different series. We expect that they can exhibit variance about translationese.
\begin{itemize}
  \item Google-Translate
  \item DeepSeek-R1-Distill-Qwen-32B~\citep{deepseekai-2025-dpskr1}
  \item DeepSeek-R1~\citep{deepseekai-2025-dpskr1}
  \item Llama3.3-70B-Instruct	
  \item Qwen2.5-3B-Instruct
  \item Qwen2.5-7B-Instruct
  \item Qwen2.5-72B-Instruct
\end{itemize}

We use the following five LLMs to produce translations in the section where T-index is compared with existing QE metrics. These models are from the same series. Thus, the results of QE can be more comparable among these models. For each experiment condition, we have 1,000 sources, 1,000 references, and 5,000 translations.

\begin{itemize}
  \item Qwen2.5-0.5B-Instruct
  \item Qwen2.5-3B-Instruct
  \item Qwen2.5-7B-Instruct
  \item Qwen2.5-32B-Instruct
  \item QwQ-32B
\end{itemize}

\section{More details about baselines}
\label{sec:further-discussion-binary}

We introduce the high-level intuition of each unsupervised baseline (when the scoring model is trained on high-translationese). 
\begin{itemize}[leftmargin=*]
    \setlength{\itemsep}{0pt}
    \item \textbf{Log-likelihood}: LMs assign higher probabilities to samples close to the training distribution. The likelihood of low-translationese samples is expected to be lower than that of high-translationese ones.
    \item \textbf{Entropy}: LMs are less uncertain about in-distribution samples, so the entropy of high-translationese samples are expected to be lower than that of low-translationese ones.
    \item \textbf{Fast-DetectGPT}~\citep{bao-etal-2024-fast}: FDG assumes that the likelihood of the original continuation after the context should be higher than that of the alternatives for machine-generated texts, distinguishing them from human-written samples. Similarly, the likelihood of low-translationese translations changes more significantly than that of high-translationese samples under substitution.
    \item \textbf{Mahalanobis Distance}~\citep{ren-etal-2023-outofdistribution}: MD measures the distance between the last hidden states of the sample and the training distribution. High-translationese samples are closer to the training distribution than low-translationese translations.
    \item \textbf{Relative Mahalanobis Distance}~\citep{ren-etal-2023-outofdistribution}: RMD provides a background distribution based on MD. OOD samples are expected to be closer to the background distribution, but get away from the training distribution. Here, we measure the relative distance of a sample to the low-translationese distribution and high-translationese distribution.
    \item \textbf{Trajectory Volatility}~\citep{wang-etal-2024-embedding}: TV measures the changes between adjacent layers of hidden states of model output when the last hidden states of the outputs cluster in a high-density region, which is observed on OOD samples in mathematical reasoning. Here, we expect that low-translationese samples cluster more closely.
    % \item \textbf{Likelihood Ratios}~\citep{ren-etal-2019-likelihood}: LLR cancels out irrelevant distribution shifts by the differences between two log-likelihoods of a sample given by two models. In our setting, both the foreignization model and the domestication model are used for scoring.
\end{itemize}

Here is the intuition of supervised baselines:

\begin{itemize}[leftmargin=*]
    \setlength{\itemsep}{0pt}
    \item \textbf{DPO-aligned}~\citep{rafailov-etal-2023-direct}: Using DPO to align an LLM to prefer high-translationese translation but penalize the low-translationese. The aligned model will assign higher probabilities to high-translationese translations than low-translationese ones.
    \item \textbf{Bradley-Terry RM}~\citep{bradley-terry-1952-rank,ouyang-etal-2022-training}: Training a reward model (RM) with Bradley-Terry loss to assign higher scores to high-translationese samples and lower scores to low-translationese.
    \item \textbf{XLM-RoBERTa}~\citep{conneau-etal-2020-unsupervised}: Fine-tuning a pre-trained encoder for classification.
    \item \textbf{SVM with linguistic features}~\citep{hu-and-kübler-2021-translated}: Extracting linguistic features and using SVM for classification.
\end{itemize}

All scoring models are trained on 1 or 2 A100-80G GPUs. 
It takes around 5 minutes to train a \texttt{Qwen2.5-0.5B} base model with the objectives including SFT, DPO, and RM. 
The implementations and hyperparameters for our model training can be found in our GitHub repository: \url{https://github.com/yikang0131/TranslationeseIndex}.

\section{Instruction for annotations and prompts used in LLM-as-a-judge}
\label{sec:anno-judge}

\begin{CJK*}{UTF8}{gbsn}
{\small
\begin{table*}[htbp]
\centering
\caption{Guidelines and instructions for human annotation}
\begin{tabular}{p{0.9\textwidth}}
\toprule
一、任务说明 [Task Description] \\
• 标注人员需要对50条机器翻译数据进行评估 [Annotators need to evaluate 50 machine translation examples] \\
• 每条数据包含一条英文原文和一条中文译文 [Each example contains an English source text and a Chinese translation] \\
• 使用0-5的Likert量表对译文的翻译腔程度进行打分 [Use a 0-5 Likert scale to rate the degree of translationese] \\
• 标注者需要从译文中摘选0-3个翻译腔严重的片段作为评分依据 [Annotators should select 0-3 segments with severe translationese as evidence] \\
• 标注数据将开源但仅用做学术用途，标注者信息会做匿名处理 [The annotated data will be open-sourced only for academic purposes, and all information of annotators will be anonymized] \\
\midrule
二、翻译腔的定义 [Definition of Translationese] \\
译文用词和语序明显受到翻译过程的影响，导致不符合原生汉语的使用习惯。主要体现在: [Word choice and word order are clearly influenced by the translation process, resulting in expressions that deviate from native Chinese usage. This mainly manifests in:] \\
1. 用词和语序造成的不自然 [Unnatural word choice and word order] \\
2. 过于直译造成的语义错误（其他类型的翻译错误不计入翻译腔评分）[Semantic errors caused by overly literal translation (other types of translation errors are not counted)] \\
\midrule
三、评分标准与摘选要求 [Rating Criteria and Selection Requirements] \\
0分：完全符合中文表达习惯，读起来自然流畅 [0 points: Completely conforms to Chinese expression habits, reads naturally and fluently] \\
1分：稍有不自然，但不影响理解 [1 point: Slightly unnatural but does not affect understanding] \\
2分：个别用词或语序有翻译腔，整体基本可接受 [2 points: Individual words or word order show translationese, but overall acceptable] \\
3分：明显的翻译腔，但基本意思清晰 [3 points: Obvious translationese, but basic meaning is clear] \\
4分：较重的翻译腔，部分表达不符合中文习惯 [4 points: Heavy translationese, some expressions deviate from Chinese conventions] \\
5分：严重的翻译腔，直译痕迹明显或造成理解障碍 [5 points: Severe translationese, obvious literal translation or causes comprehension difficulties] \\
\midrule
四、示例分析 [Example Analysis] \\
示例1 [Example 1]：\\
原文 [Source]: This issue requires immediate attention. \\
译文 [Translation]: 这个议题需要即刻的关注。\\
评分 [Rating]: 3分 \\
翻译腔片段 [Translationese segments]："即刻的关注" \\
\midrule
五、标注注意事项 [Annotation Notes] \\
1. 评分时应重点关注译文的流畅度和自然程度 [Focus on fluency and naturalness when rating] \\
2. 摘选片段时应选择最能体现翻译腔的部分 [Select segments that best demonstrate translationese] \\
3. 摘选数量应与翻译腔程度相匹配 [Number of selections should match the degree of translationese] \\
4. 不要将其他类型的翻译错误记入翻译腔评分 [Do not count other types of translation errors in translationese rating] \\
\bottomrule
\end{tabular}
\end{table*}
}

{\small
\begin{table*}[htbp]
\centering
\caption{Prompts used in the LLM-as-a-judge method}
\begin{tabular}{p{0.9\textwidth}}
\toprule
【LLM-as-a-Judge: 中文翻译质量评估与解析】\\

任务描述：
在以下提供的两个英文句子的中文翻译版本中（A和B），请判断哪一个翻译更符合中文的表达习惯和语境。同时，请简要说明选择的理由。最后，您的回答应包含“A”或“B”，以表明您认为哪一版本更优。
\\
示例：
\\
英文原句：The quick brown fox jumps over the lazy dog. \\
翻译A：那只敏捷的棕色狐狸跳过了懒惰的狗。 \\
翻译B：快速的棕色狐狸跃过懒散的狗。 \\

评估解析：翻译A使用了“敏捷”和“懒惰”这两个形容词，更加形象生动，更符合中文表达习惯中的具体性和形象性。而“跃过”相较于“跳过”在中文中更具有画面感。因此，翻译A更优。 \\
评估结果：A \\

实际任务： \\

英文原文1：\{\{source\}\} \\
翻译A：\{\{translation\_A\}\} \\
翻译B：\{\{translation\_B\}\} \\
\bottomrule
\end{tabular}
\end{table*}
}

\end{CJK*}
\end{document}